\documentclass[10pt,twocolumn,letterpaper]{article}

\usepackage{cvpr}
\usepackage{times}
\usepackage{epsfig}
\usepackage{graphicx}
\usepackage{amsmath}
\usepackage{amssymb}
\usepackage{booktabs}
\usepackage{tabularx}

\usepackage{xcolor}
\definecolor{light-gray}{gray}{0.95}

\usepackage[pagebackref=true,breaklinks=true,letterpaper=true,colorlinks,bookmarks=false]{hyperref}

\cvprfinalcopy 

\def\cvprPaperID{6160} 

\ifcvprfinal\fi
\begin{document}

\title{High Quality Monocular Depth Estimation via Transfer Learning}

\author{Ibraheem Alhashim\\
KAUST\\
{\tt\small ibraheem.alhashim@kaust.edu.sa}
\and
Peter Wonka\\
KAUST\\
{\tt\small pwonka@gmail.com}
}

\maketitle

\begin{abstract}
    Accurate depth estimation from images is a fundamental task in many applications including scene understanding and reconstruction. Existing solutions for depth estimation often produce blurry approximations of low resolution.
    This paper presents a convolutional neural network for computing a high-resolution depth map given a single RGB image with the help of transfer learning. Following a standard encoder-decoder architecture, we leverage features extracted using high performing pre-trained networks when initializing our encoder along with augmentation and training strategies that lead to more accurate results. We show how, even for a very simple decoder, our method is able to achieve detailed high-resolution depth maps. Our network, with fewer parameters and training iterations, outperforms state-of-the-art on two datasets and also produces qualitatively better results that capture object boundaries more faithfully. Code and corresponding pre-trained weights are made publicly available\footnote{https://github.com/ialhashim/DenseDepth}.
\end{abstract}

\section{Introduction}

Depth estimation from 2D images is a fundamental task in many applications including scene understanding and reconstruction \cite{Lee2011,moreno2007active,Hazirbas2016FuseNetID}. Having a dense depth map of the real-world can be very useful in applications including navigation and scene understanding, augmented reality \cite{Lee2011}, image refocusing \cite{moreno2007active}, and segmentation \cite{Hazirbas2016FuseNetID}. Recent developments in depth estimation are focusing on using convolutional neural networks (CNNs) to perform 2D to 3D reconstruction.
While the performance of these methods has been steadily increasing, there are still major problems in both the quality and the resolution of these estimated depth maps.
Recent applications in augmented reality, synthetic depth-of-field, and other image effects \cite{Hedman2018,Cao2018,Wang2018} require fast computation of high resolution 3D reconstructions in order to be applicable. For such applications, it is critical to faithfully reconstruct discontinuity in the depth maps and avoid the large perturbations that are often present in depth estimations computed using current CNNs.

\begin{figure}[t]
\begin{center}
\includegraphics[width=\linewidth]{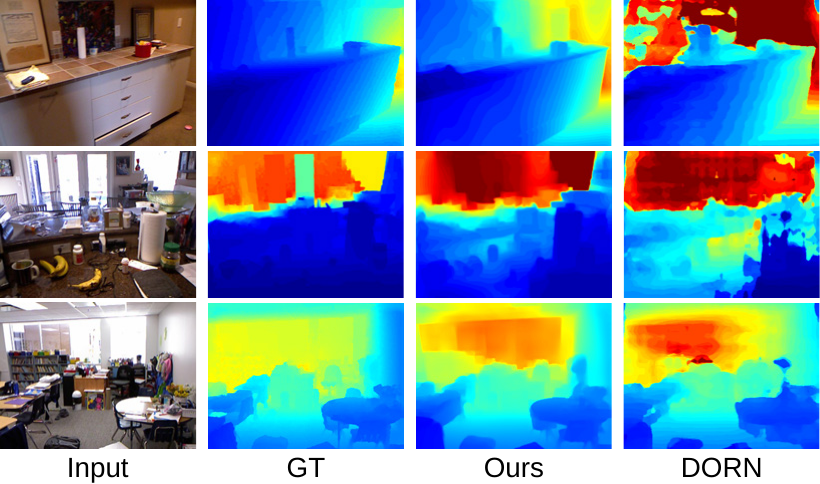}
\end{center}
   \caption{\textbf{Comparison of estimated depth maps:} input RGB images, ground truth depth maps, our estimated depth maps, state-of-the-art results of \cite{Fu2018DeepOR}.}
\label{fig:teaser}
\end{figure}

Based on our experimental analysis of existing architectures and training strategies \cite{Eigen2014,Li2015,Laina2016,Xu2017,Fu2018DeepOR} we set out with the design goal to develop a simpler architecture that makes training and future modifications easier. Despite, or maybe even due to its simplicity, our architecture produces depth map estimates of higher accuracy and significantly higher visual quality than those generated by existing methods (see Fig.~\ref{fig:teaser}). To achieve this, we rely on transfer learning were we repurpose high performing pre-trained networks that are originally designed for image classification as our deep features encoder. A key advantage of such a transfer learning-based approach is that it allows for a more modular architecture where future advances in one domain are easily transferred to the depth estimation problem.

\begin{figure*}
\begin{center}
\includegraphics[width=\linewidth]{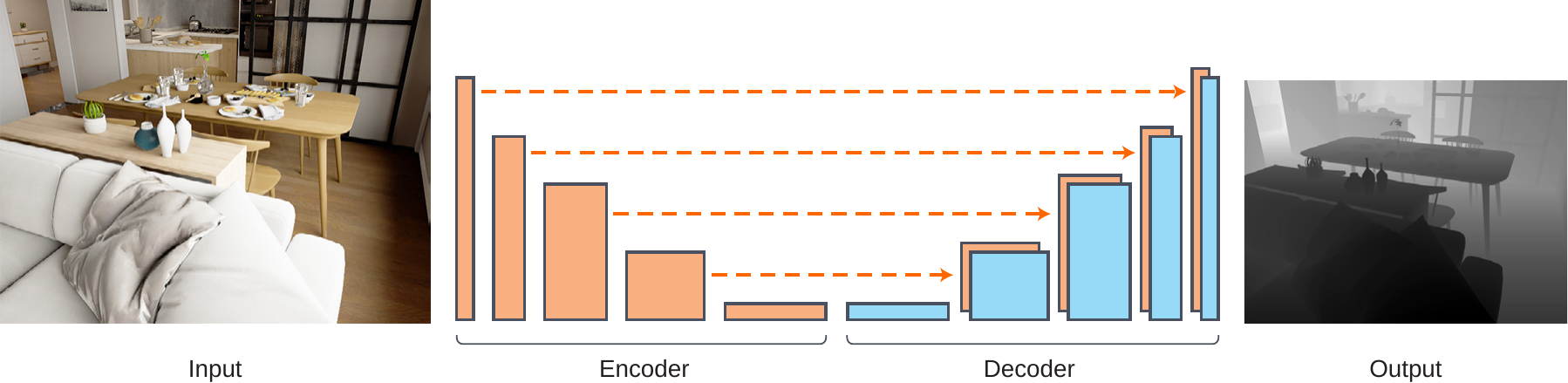}
\end{center}
   \caption{\textbf{Overview of our network architecture.} We employ a straightforward encoder-decoder architecture with skip connections. The encoder part is a pre-trained truncated DenseNet-169 \cite{huang2017densely} with no additional modifications. The decoder is composed of basic blocks of convolutional layers applied on the concatenation of the $2\times$ bilinear upsampling of the previous block with the block in the encoder with the same spatial size after upsampling. }
\label{fig:network_overview}
\end{figure*}

\paragraph{Contributions:} Our contributions are threefold. First, we propose a simple transfer learning-based network architecture that produces depth estimations of higher accuracy and quality. The resulting depth maps capture object boundaries more faithfully than those generated by existing methods with fewer parameters and less training iterations. Second, we define a corresponding loss function, learning strategy, and simple data augmentation policy that enable faster learning. Third, we propose a new testing dataset of photo-realistic synthetic indoor scenes, with perfect ground truth, to better evaluate the generalization performance of depth estimating CNNs.

We perform different experiments on several datasets to evaluate the performance and quality of our depth estimating network. The results show that our approach not only outperforms the state-of-the-art and produces high quality depth maps on standard depth estimation datasets, but it also results in the best generalization performance when applied to a novel dataset.

\section{Related Work}

The problem of 3D scene reconstruction from RGB images is an ill-posed problem. Issues such as lack of scene coverage, scale ambiguities, translucent or reflective materials all contribute to ambiguous cases where geometry cannot be derived from appearance. In practice, the more successful approaches for capturing a scene's depth rely on hardware assistance, e.g. using laser or IR-based sensors, or require a large number of views captured using high quality cameras followed by a long and expensive offline reconstruction process. Recently, methods that rely on CNNs are able to produce reasonable depth maps from a single or couple of RGB input images at real-time speeds. In the following, we look into some of the works that are relevant to the problem of depth estimation and 3D reconstruction from RGB input images. More specifically, we look into recent solutions that depend on deep neural networks.

\paragraph{Monocular depth estimation} has been considered by many CNN methods where they formulate the problem as a regression of the depth map from a single RGB image \cite{Eigen2014,Laina2016,Xu2017,Hao2018DetailPD,Xu2018StructuredAG,Fu2018DeepOR}. While the performance of these methods have been increasing steadily, general problems in both the quality and resolution of the estimated depth maps leave a lot of room for improvement. Our main focus in this paper is to push towards generating higher quality depth maps with more accurate boundaries using standard neural network architectures. Our preliminary results do indicate that improvements on the state-of-the-art are possible to achieve by leveraging existing simple architectures that perform well on other computer vision tasks.

\paragraph{Multi-view} stereo reconstruction using CNN algorithms have been recently proposed \cite{Huang2018DeepMVSLM}. Prior work considered the subproblem that looks at image pairs \cite{Ummenhofer2017}, or three consecutive frames \cite{Godard2018DiggingIS}. Joint key-frame based dense camera tracking and depth map estimation was presented by \cite{Zhou2018DeepTAMDT}. In this work, we seek to push the performance for single image depth estimation. We suspect that the features extracted by monocular depth estimators could also help derive better multi-view stereo reconstruction methods.

\paragraph{Transfer learning} approaches have been shown to be very helpful in many different contexts. In recent work, Zamir et al. investigated the efficiency of transfer learning between different tasks~\cite{Zamir2018TaskonomyDT}, many of which were are related to 3D reconstruction. Our method is heavily based on the idea of transfer learning where we make use of image encoders originally designed for the problem of image classification \cite{huang2017densely}. We found that using such encoders that do not aggressively downsample the spatial resolution of the input tend to produce sharper depth estimations especially with the presence of skip connections.

\paragraph{Encoder-decoder} networks have made significant contributions in many vision related problems such as image segmentation \cite{Ronneberger2015u}, optical flow estimation \cite{Dosovitskiy2015}, and image restoration \cite{LehtinenMHLKAA18}. In recent years, the use of such architectures have shown great success both in the supervised and the unsupervised setting of the depth estimation problem \cite{Godard2017,Ummenhofer2017,Huang2018DeepMVSLM,Zhou2018DeepTAMDT}. Such methods typically use one or more encoder-decoder network as a sub part of their larger network. In this work, we employ a single straightforward encoder-decoder architecture with skip connections (see Fig. \ref{fig:network_overview}). Our results indicate that it is possible to achieve state-of-the-art high quality depth maps using a simple encoder-decoder architecture.

 \section{Proposed Method} \label{sec:method}

In this section, we describe our method for estimating a depth map from a single RGB image. We first describe the employed encoder-decoder architecture. We then discuss our observations on the complexity of both encoder and decoder and its relation to performance. Next, we propose an appropriate loss function for the given task. Finally, we describe efficient augmentation policies that help the training process significantly.

\subsection{Network Architecture}

\paragraph{Architecture.} Fig. \ref{fig:network_overview} shows an overview of our encoder-decoder network for depth estimation. For our \textit{encoder}, the input RGB image is encoded into a feature vector using the DenseNet-169 \cite{huang2017densely} network pretrained on ImageNet \cite{Deng2009}. This vector is then fed to a successive series of up-sampling layers \cite{LehtinenMHLKAA18}, in order to construct the final depth map at half the input resolution. These upsampling layers and their associated skip-connections form our \textit{decoder}. Our decoder does not contain any Batch Normalization \cite{Ioffe2015BNA} or other advanced layers recommended in recent state-of-the-art methods \cite{Fu2018DeepOR,Hao2018DetailPD}. Further details about the architecture and its layers along with their exact shapes are described in the appendix.

\paragraph{Complexity and performance.} The high performance of our surprisingly simple architecture gives rise to questions about which components contribute the most towards achieving these quality depth maps. We have experimented with different state-of-the-art encoders \cite{Bianco2018}, of more or less complexity than that of DenseNet-169, and we also looked at different decoder types \cite{Laina2016, Wojna2017TheDI}. What we experimentally found is that, in the setting of an encoder-decoder architecture for depth estimation, recent trends of having convolutional blocks exhibiting more complexity do not necessarily help the performance. This leads us to advocate for a more thorough investigation when adopting such complex components and architectures. Our experiments show that a simple decoder made of a $2\times$ bilinear upsampling step followed by two standard convolutional layers performs very well.

\begin{figure*}[t]
\begin{center}
\includegraphics[width=\linewidth]{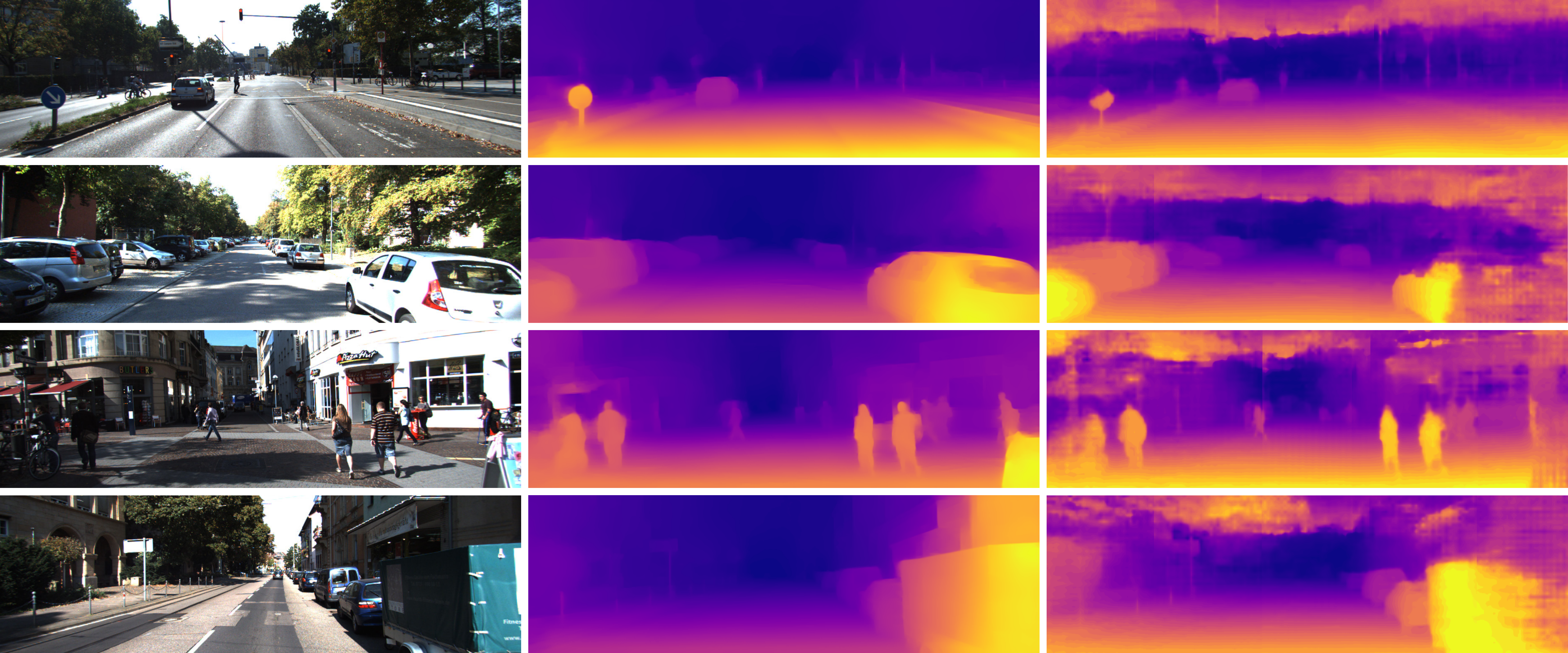}
\end{center}
   \caption{\textbf{Qualitative results from the KITTI dataset:} input RGB images, our estimated depth maps, state-of-the-art results of \cite{Fu2018DeepOR}.}
\label{fig:kitti}
\end{figure*}

\subsection{Learning and Inference}

\paragraph{Loss Function.} A standard loss function for depth regression problems considers the difference between the ground-truth depth map $y$ and the prediction of the depth regression network $\hat{y}$ \cite{Eigen2014}. Different considerations regarding the loss function can have a significant effect on the training speed and the overall depth estimation performance. Many variations on the loss function employed for optimizing the neural network can be found in the depth estimation literature \cite{Eigen2014,Laina2016,Ummenhofer2017,Fu2018DeepOR}. In our method, we seek to define a loss function that balances between reconstructing depth images by minimizing the difference of the depth values while also penalizing distortions of high frequency details in the image domain of the depth map. These details typically correspond to the boundaries of objects in the scene.

For training our network, we define the loss $L$ between $y$ and $\hat{y}$ as the weighted sum of three loss functions: 
\begin{equation}
L(y,\hat{y}) = \lambda L_{depth}(y,\hat{y}) + L_{grad}(y,\hat{y}) + L_{SSIM}(y,\hat{y}).
\end{equation}

The first loss term $L_{depth}$ is the point-wise L1 loss defined on the depth values:
\begin{equation}
L_{depth}(y,\hat{y}) = \frac{1}{n} \sum_{p}^{n} \lvert y_p -\hat{y}_p \rvert.
\end{equation}

The second loss term $L_{grad}$ is the L1 loss defined over the image gradient $\boldsymbol{g}$ of the depth image:
\begin{equation}
L_{grad}(y,\hat{y}) = \frac{1}{n} \sum_{p}^{n} \lvert \boldsymbol{g_\mathrm{x}}(y_p,\hat{y}_p) \rvert +                         \lvert \boldsymbol{g_\mathrm{y}}(y_p,\hat{y}_p) \rvert
\end{equation}
where $\boldsymbol{g_\mathrm{x}}$ and $\boldsymbol{g_\mathrm{y}}$, respectively, compute the differences in the $\mathrm{x}$ and $\mathrm{y}$ components for the depth image gradients of $y$ and $\hat{y}$. 

Lastly, $L_{SSIM}$ uses the Structural Similarity (SSIM) \cite{Wang2004SSIM} term which is a commonly-used metric for image reconstruction tasks. It has been recently shown to be a good loss term for depth estimating CNNs \cite{Godard2017}. Since SSIM has an upper bound of one, we define it as a loss $L_{SSIM}$ as follows:
\begin{equation}
L_{SSIM}(y,\hat{y}) = \frac{1 - SSIM(y,\hat{y})}{2}.
\end{equation}
Note that we only define one weight parameter $\lambda$ for the loss term $L_{depth}$. We empirically found and set $\lambda=0.1$ as a reasonable weight for this term. 

An inherit problem with such loss terms is that they tend to be larger when the ground-truth depth values are bigger. In order to compensate for this issue, we consider the reciprocal of the depth \cite{Ummenhofer2017, Huang2018DeepMVSLM} where for the original depth map $y_{orig}$ we define the target depth map $y$ as $y = m / y_{orig}$ where $m$ is the maximum depth in the scene (e.g. $m=10$ meters for the NYU Depth v2 dataset). Other methods consider transforming the depth values and computing the loss in the log space \cite{Eigen2014,Ummenhofer2017}.

\paragraph{Augmentation Policy.} Data augmentation, by geometric and photo-metric transformations, is a standard practice to reduce over-fitting leading to better generalization performance \cite{krizhevsky2012imagenet}.
Since our network is designed to estimate depth maps of an entire image, not all geometric transformations would be appropriate since distortions in the image domain do not always have meaningful geometric interpretations on the ground-truth depth. Applying a vertical flip to an image capturing an indoor scene may not contribute to the learning of expected statistical properties (e.g. geometry of the floors and ceilings). Therefore, we only consider horizontal flipping (i.e. mirroring) of images at a probability of $0.5$. Image rotation is another useful augmentation strategy, however, since it introduces invalid data for the corresponding ground-truth depth we do not include it. 
For photo-metric transformations we found that applying different color channel permutations, e.g. swapping the red and green channels on the input, results in increased performance while also being extremely efficient. We set the probability for this color channel augmentation to $0.25$. Finding improved data augmentation policies and their probability values for the problem of depth estimation is an interesting topic for future work \cite{Cubuk2018AutoAugmentLA}.


\section{Experimental Results}

In this section we describe our experimental results and compare the performance of our network to existing state-of-the-art methods. Furthermore, we perform ablation studies to analyze the influence of the different parts of our proposed method. Finally, we compare our results on a newly proposed dataset of high quality depth maps in order to better test the generalization and robustness of our trained model.

\subsection{Datasets}

\paragraph{NYU Depth v2} is a dataset that provides images and depth maps for different indoor scenes captured at a resolution of $640\times480$ \cite{Silberman2012}. The dataset contains 120K training samples and 654 testing samples \cite{Eigen2014}. We train our method on a 50K subset. Missing depth values are filled using the inpainting method of \cite{Levin2004}. The depth maps have an upper bound of 10 meters. Our network produces predictions at half the input resolution, i.e. a resolution of $320\times240$. For training, we take the input images at their original resolution and downsample the ground truth depths to $320\times240$. Note that we do not crop any of the input image-depth map pairs even though they contain missing pixels due to a distortion correction preprocessing. During test time, we compute the depth map prediction of the full test image and then upsample it by $2\times$ to match the ground truth resolution and evaluate on the pre-defined center cropping by Eigen et al. \cite{Eigen2014}. At test time, we compute the final output by taking the average of an image's prediction and the prediction of its mirror image.

\paragraph{KITTI} is a dataset that provides stereo images and corresponding 3D laser scans of outdoor scenes captured using equipment mounted on a moving vehicle \cite{geiger2013vision}. The RGB images have a resolution of around $1241\times376$ while the corresponding depth maps are of very low density with lots of missing data. We train our method on a subset of around 26K images, from the left view, corresponding to scenes not included in the 697 test set specified by \cite{Eigen2014}. Missing depth values are filled using the inpainting method mentioned earlier. The depth maps have an upper bound of 80 meters. Our encoder's architecture expects image dimensions to be divisible by 32 \cite{huang2017densely}, therefore, we upsample images bilinearly to $1280\times384$ during training. During testing, we first scale the input image to the expected resolution and then upsample the output depth image from $624\times192$ to the original input resolution. The final output is computed by taking the average of an image's prediction and the prediction of its mirror image.

\begin{figure*}[t]
\begin{center}
\includegraphics[width=\linewidth]{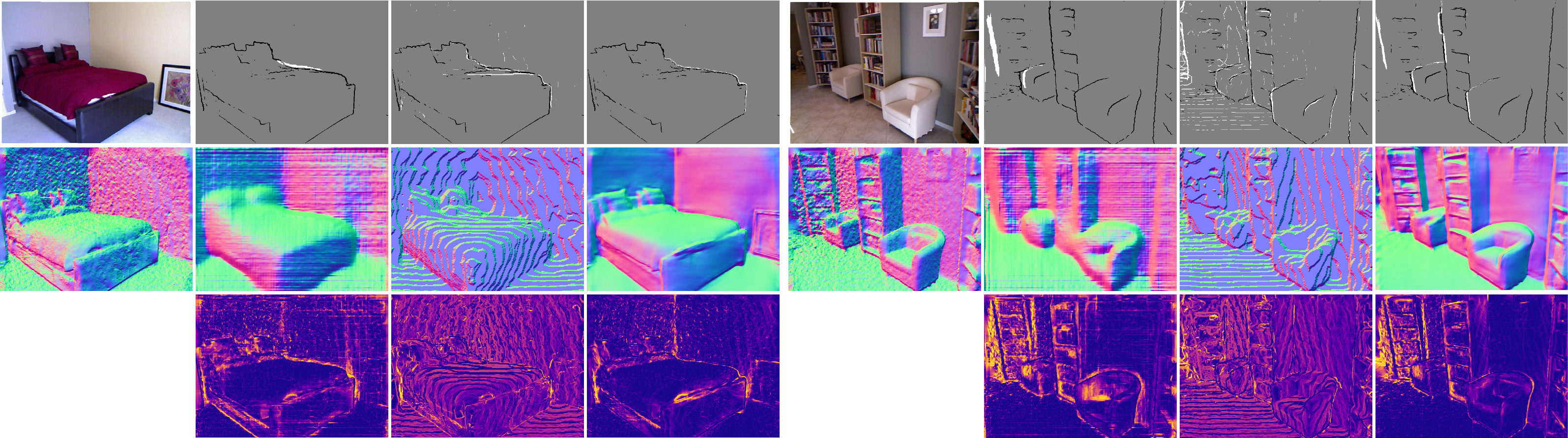}
\end{center}
   \caption{\textbf{Qualitative measures.} The left most column shows the input image (top) and its extracted normal map (bottom) using the ground truth depth. For the following columns, the top row visualizes the difference in the thresholded gradient magnitude image of the estimated depths computed using Laina et al. \cite{Laina2016}, Fu et al. \cite{Fu2018DeepOR}, and our method. Bright regions represent false edges while dark regions are remaining missed edges. The middle row shows the corresponding extracted normal maps. The bottom row visualizes the surface normal error. Note that since the method of \cite{Fu2018DeepOR} generates depth maps with sharp steps, computing a reasonable normal map is not straightforward. }
\label{fig:qualitative}
\end{figure*}

\begin{table*}[t]
\centering
\begin{tabular}{l|lll|lll}
\toprule
Method         & $\delta_{1}\uparrow$ & $\delta_{2}\uparrow$ & $\delta_{3}\uparrow$ & rel$\downarrow$   & rms$\downarrow$   & $log_{10}\downarrow$  \\ 
\midrule
Eigen et al. \cite{Eigen2014}    & 0.769  & 0.950  & 0.988 & 0.158 & 0.641 & -   \\
Laina et al. \cite{Laina2016}    & 0.811  & 0.953  & 0.988 & 0.127 & 0.573 & 0.055   \\
MS-CRF \cite{Xu2017}             & 0.811  & 0.954  & 0.987 & 0.121 & 0.586 & 0.052   \\
Hao et al. \cite{Hao2018DetailPD}& 0.841  & 0.966  & 0.991 & 0.127 & 0.555 & 0.053   \\
Fu et al. \cite{Fu2018DeepOR}    & 0.828  & 0.965  & 0.992 & \textbf{0.115} & 0.509 & \textbf{0.051}   \\
Ours                   & \textbf{0.846}  &  \textbf{0.974}  &  \textbf{0.994} &  0.123  &  \textbf{0.465} &  0.053   \\
\midrule
Ours (scaled)      & \textbf{0.895}  &  \textbf{0.980}  &  \textbf{0.996} &  \textbf{0.103}  &  \textbf{0.390} &  \textbf{0.043}   \\
\bottomrule
\end{tabular}
\bigskip
\caption{\textbf{Comparisons of different methods on the NYU Depth v2 dataset.} The reported numbers are from the corresponding original papers. The last row shows results obtained using our method with applied scaling that matches the median with the ground truth \cite{Zhou2017}. }
\label{tab:1}
\end{table*}

\begin{table*}[t]
\centering
\begin{tabular}{l|lll|llll}
\toprule
Method         & $\delta_{1}\uparrow$ & $\delta_{2}\uparrow$ & $\delta_{3}\uparrow$ & rel$\downarrow$ & sq. rel$\downarrow$    & rms$\downarrow$   & $log_{10}\downarrow$  \\ 
\midrule
Eigen et al. \cite{Eigen2014}       & 0.692  & 0.899  & 0.967 & 0.190 & 1.515 & 7.156 & 0.270 \\
Godard et al. \cite{Godard2017}      & 0.861  & 0.949  & 0.976 & 0.114 & 0.898 & 4.935  & 0.206 \\
Kuznietsov et al. \cite{Kuznietsov2017}  & 0.862  & 0.960  & 0.986 & 0.113 & 0.741 & 4.621 & 0.189 \\
Fu et al. \cite{Fu2018DeepOR}    & \textbf{0.932}  & \textbf{0.984}  & \textbf{0.994} & \textbf{0.072} & \textbf{0.307} & \textbf{2.727}  & \textbf{0.120} \\
Ours                             & \underline{0.886}  & \underline{0.965}  & \underline{0.986} & \underline{0.093} & \underline{0.589} & \underline{4.170} & \underline{0.171} \\
\bottomrule
\end{tabular}
\bigskip
\caption{\textbf{KITTI dataset.} We compare our method against the state-of-the-art on this dataset. Measurements are made for the depth range from $0m$ to $80m$. The best results are bolded, and the second best are underlined.}
\label{tab:kitti}
\end{table*}

\subsection{Implementation Details} 

We implemented our proposed depth estimation network using TensorFlow \cite{tensorflow2015-whitepaper} and trained on four NVIDIA TITAN Xp GPUs with 12GB memory. Our encoder is a DenseNet-169 \cite{huang2017densely} pretrained on ImageNet \cite{Deng2009}. The weights for the decoder are randomly initialized following \cite{glorot2010understanding}. In all experiments, we used the ADAM \cite{jlb2015adam} optimizer with learning rate $0.0001$ and parameter values $\beta_1=0.9$, $\beta_2=0.999$. The batch size is set to 8. The total number of trainable parameters for the entire network is approximately 42.6M parameters. Training is performed for 1M iterations for NYU Depth v2, needing 20 hours to finish. Training for the KITTI dataset is performed for 300K iterations, needing 9 hours to train.

\subsection{Evaluation}

\paragraph{Quantitative evaluation.} We quantitatively compare our method against state-of-the-art using the standard six metrics used in prior work \cite{Eigen2014}. These error metrics are defined as:
\begin{itemize}
    \item average relative error (rel): $\frac{1}{n}\sum_p^n \frac{\lvert y_p-\hat{y}_p \rvert}{y}$;
    \item root mean squared error (rms): $\sqrt{\frac{1}{n}\sum_p^n (y_p-\hat{y}_p)^2)}$;
    \item average ($\log_{10}$) error: $\frac{1}{n}\sum_p^n \lvert \log_{10}(y_p)-\log_{10}(\hat{y}_p) \rvert$;
    \item threshold accuracy ($\delta_i$): $\%$ of $y_p$ s.t. $\text{max}(\frac{y_p}{\hat{y}_p},\frac{\hat{y}_p}{y_p}) = \delta < thr$ for $thr=1.25,1.25^2,1.25^3$;
\end{itemize}
where $y_p$ is a pixel in depth image $y$, $\hat{y}_p$ is a pixel in the predicted depth image $\hat{y}$, and $n$ is the total number of pixels for each depth image.

\paragraph{Qualitative results.} We conduct three experiments to approximately evaluate the quality of the results using three measures on the NYU Depth v2 test set. The first measure is a perception-based qualitative metric that measures the quality of the results by looking at the similarity of the resulting depth maps in image space. We do so by rendering a gray scale visualization of the ground truth and that of the predicted depth map and then we compute the mean structural similarity term (mSSIM) of the entire test dataset $\frac{1}{T} \sum_i^T {SSIM}(y_i,\hat{y}_i)$. The second measure considers the edges formed in the depth map. For each sample, we compute the gradient magnitude image of both the ground truth and the predicted depth image, using the Sobel gradient operator \cite{Sobel1968}, and then threshold this image at values greater than 0.5 and compute the F1 score averaged across the set. The third measure is the mean cosine distance between normal maps extracted from the depth images of the ground truth and the predicted depths also averaged across the set. Fig. \ref{fig:qualitative} shows visualizations of some of these measures.

Fig. \ref{fig:gallery} shows a gallery of depth estimation results that are predicated using our method along with a comparison to those generated by the state-of-the-art.
As can be seen, our approach produces depth estimations at higher quality where depth edges better match those of the ground truth and with significantly fewer artifacts. 

\begin{table}[t]
\centering
\begin{tabular}{l|lll}
\toprule
Method         & mSSIM$\uparrow$ & F1$\uparrow$ & mne$\downarrow$  \\ 
\midrule
Laina et al. \cite{Laina2016}    & 0.957     &  0.395  & 0.698  \\
Fu et al. \cite{Fu2018DeepOR}    & 0.949      &  0.351  & 0.730  \\
\textbf{Ours}                    & \textbf{0.968}  & \textbf{0.519} & \textbf{0.636} \\
\bottomrule
\end{tabular}
\bigskip
\caption{\textbf{Qualitative evaluation.} For the NYU Depth v2 testing set, we compute three measures that reflect the quality of the depth maps generated by different methods. The measures are: mean SSIM of the depth maps, mean F1 score of the edge maps, and mean of the surface normal errors. Higher values indicate better quality for the first two measures while lower values are better for the third. }
\label{tab:2}
\end{table}

\subsection{Comparing Performance}

In Tab. \ref{tab:1}, the performance of our depth estimating network is compared to the state-of-the-art on the NYU Depth v2 dataset. As can be seen, our model achieves state-of-the-art on all but two quantitative metrics. Our model is able to outperform the existing state-of-the-art \cite{Fu2018DeepOR} while requiring fewer parameters, 42.6M vs 110M, fewer number of training iterations, 1M vs 3M, and with fewer input training data, 50K samples vs 120K samples. A typical source of error for single image depth estimation networks is the estimated absolute scale of the scene. The last row in Tab. \ref{tab:1} shows that when accounting for this error, by multiplying the predicted depths by a scalar that matches the median with the ground truth \cite{Zhou2017}, we are able to achieve with a good margin state-of-the-art for the NYU Depth v2 dataset on all metrics. The results in Tab. \ref{tab:2} show that for the same dataset our method outperforms state-of-the-art on our defined quality approximating measures. We conduct these experiments for methods with published pre-trained models and code.

In Tab. \ref{tab:kitti}, the performance of our network is compared to the state-of-the-art on the KITTI dataset. Our method is the second best on all the standard metrics. We suspect that one reason our method does not outperform the state-of-the-art on this particular dataset is due to the nature of the provided depth maps. Since our loss function is designed to not only consider point-wise differences but also optimize for edges and appearance preservation by looking at regions around each point, the learning process does not converge well for very sparse depth images. Fig. \ref{fig:kitti} clearly shows that while quantitatively our method might not be the best, the quality of the produced depth maps is much better than those produced by the state-of-the-art.

\subsection{Ablation Studies}

\begin{figure}
\begin{center}
\includegraphics[width=\linewidth]{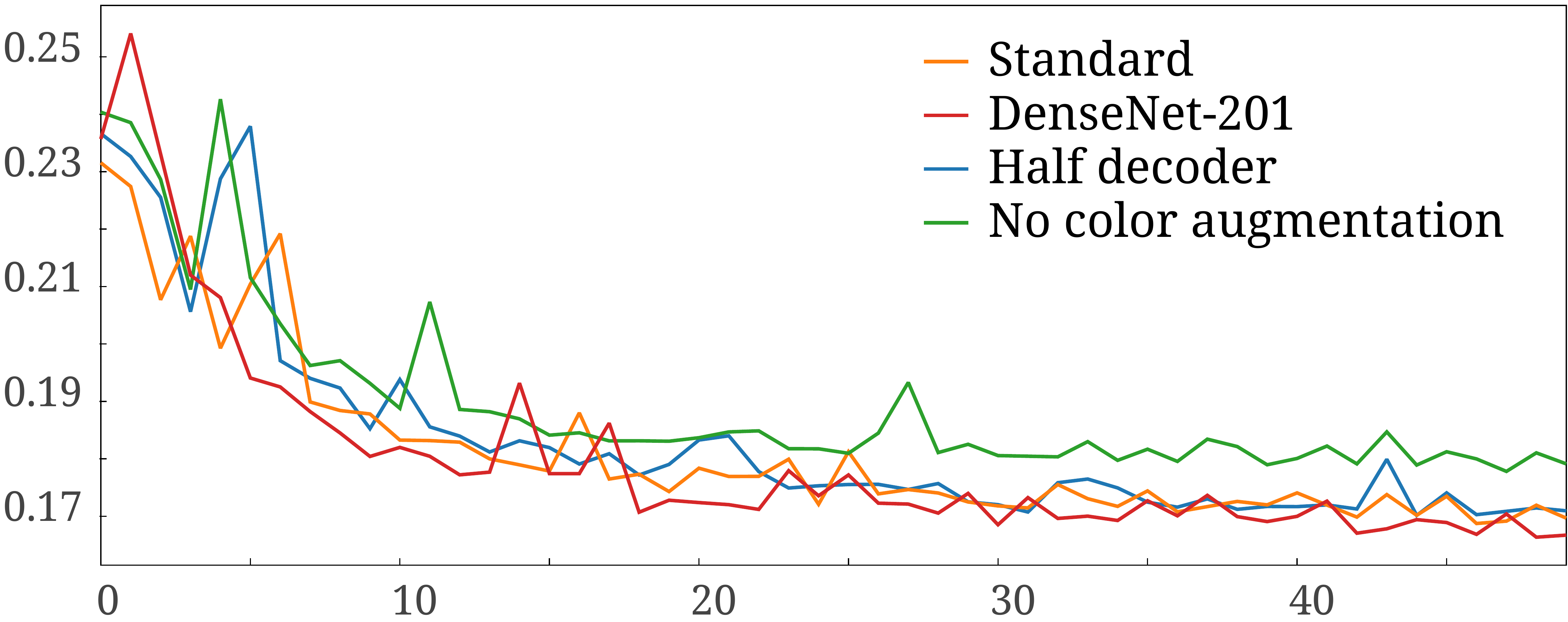}
\end{center}
   \caption{\textbf{Ablation Studies.} Three variations on our standard model are considered. \emph{DenseNet-201} (red) refers to a deeper version of the encoder. The \emph{half decoder} variation (blue) represents the model with only half the features coming out of the last layer in the encoder. Lastly, we consider the performance when disabling the \emph{color-swapping} data augmentations (green). }
\label{fig:ablation}
\end{figure}

\begin{figure*}[t]
\begin{center}
\includegraphics[width=\linewidth]{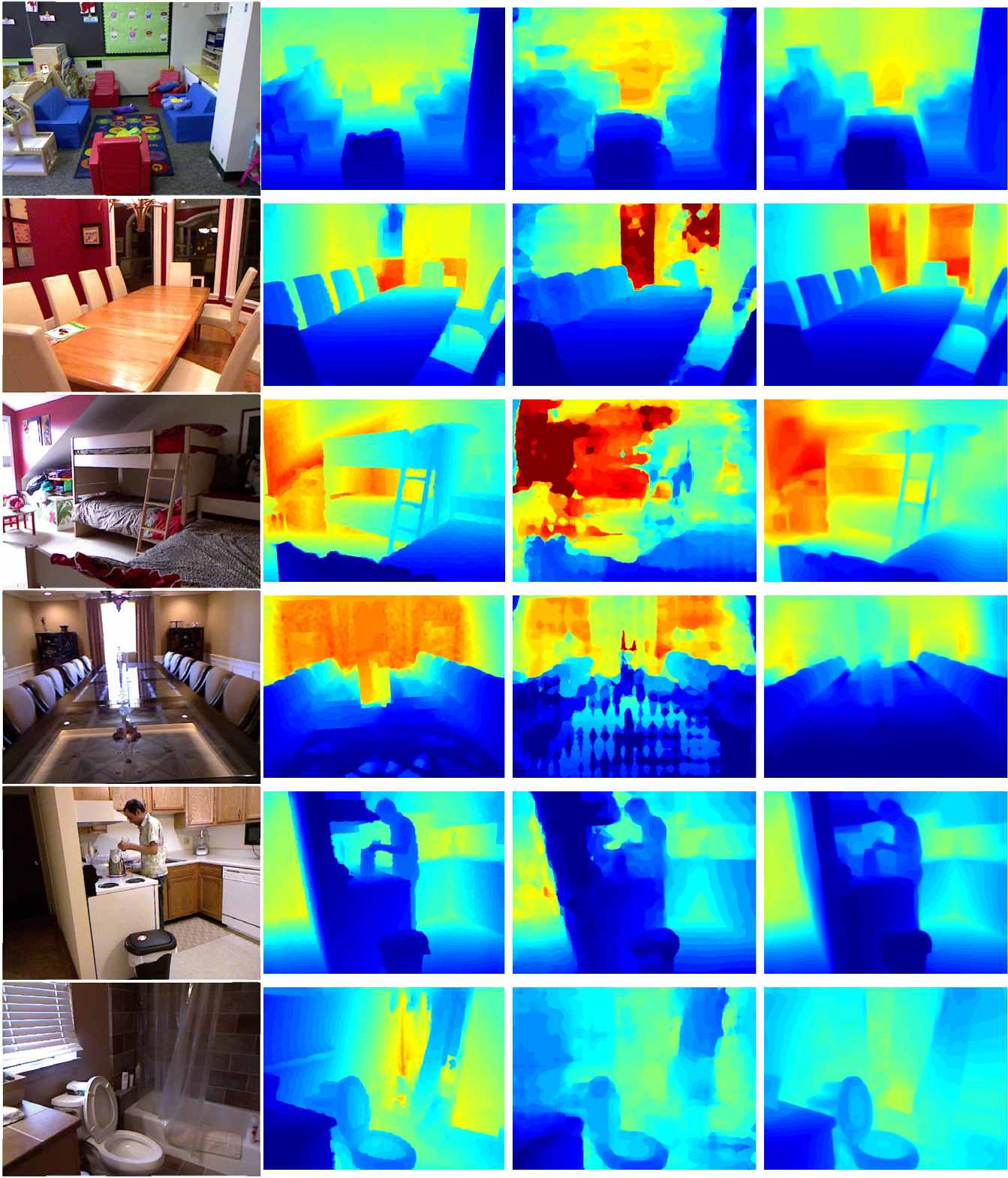}
\end{center}
   \caption{\textbf{A gallery of estimated depth maps on the NYU Depth v2 dataset:} input RGB images, ground truth depth maps, state-of-the-art results of \cite{Fu2018DeepOR} (provided by the authors), our estimated depth maps. Note that, for better visualization, we normalize all depth maps with respect to the range in its specific ground truth. }
\label{fig:gallery}
\end{figure*}

We perform ablation studies to analyze the details of our proposed architecture. Fig. \ref{fig:ablation} shows a representative look into the testing performance, in terms of validation loss, when changing some parts of our standard model or modifying our training strategy. Note that we performed these tests on a smaller subset of the NYU Depth v2 dataset.

\paragraph{Encoder depth.} In this experiment we substitute the pretrained DenseNet-169 with a denser encoder, namely the DenseNet-201. In Fig. \ref{fig:ablation} (red), we can see the validation loss is lower than that of our standard model. The big caveat, though, is that the number of parameters in the network grows by more than $2\times$. When considering using DenseNet-201 as our encoder, we found that the gains in performance did not justify the slow learning time and the extra GPU memory required.

\paragraph{Decoder depth.} In this experiment we apply a depth reducing convolution such that the features feeding into the decoder are half what they are in the standard DenseNet-169. In Fig. \ref{fig:ablation} (blue), we see a reduction in the performance and overall instability. Since these experiments are not representative of a full training session the performance difference in halving the features might not be as visible as we have observed when running full training session.

\paragraph{Color Augmentation.} In this experiment, we turn off our color channel swapping-based data augmentation. In Fig. \ref{fig:ablation} (green), we can see a significant reduction as the model tends to quickly falls into overfitting to the training data. We think this simple data augmentation and its significant effect on the neural network is an interesting topic for future work.

\subsection{Generalizing to Other Datasets}

To illustrate how well our method generalizes to other datasets, we propose a new dataset of photo-realistic indoor scenes with nearly perfect ground truth depths. These scenes are collected from the Unreal marketplace community \cite{UnrealMarket2018}. We refer to this dataset as \textbf{Unreal-1k}. It is a random sampling of 1000 images with their corresponding depth maps  selected from renderings of 32 virtual scenes using the Unreal Engine. Further details about this dataset can be found in the appendix. We compare our NYU Depth v2 trained model to two supervised methods that are also trained on the same dataset. For inference, we use the public implementations for each method. The hope of this experiment is to demonstrate how well do models trained on one dataset perform when presented with data sampled from a different distribution (i.e. synthetic vs. real, perfect depth capturing vs. a Kinect, etc.).

\begin{figure*}[t]
\begin{center}
\includegraphics[width=0.98\linewidth]{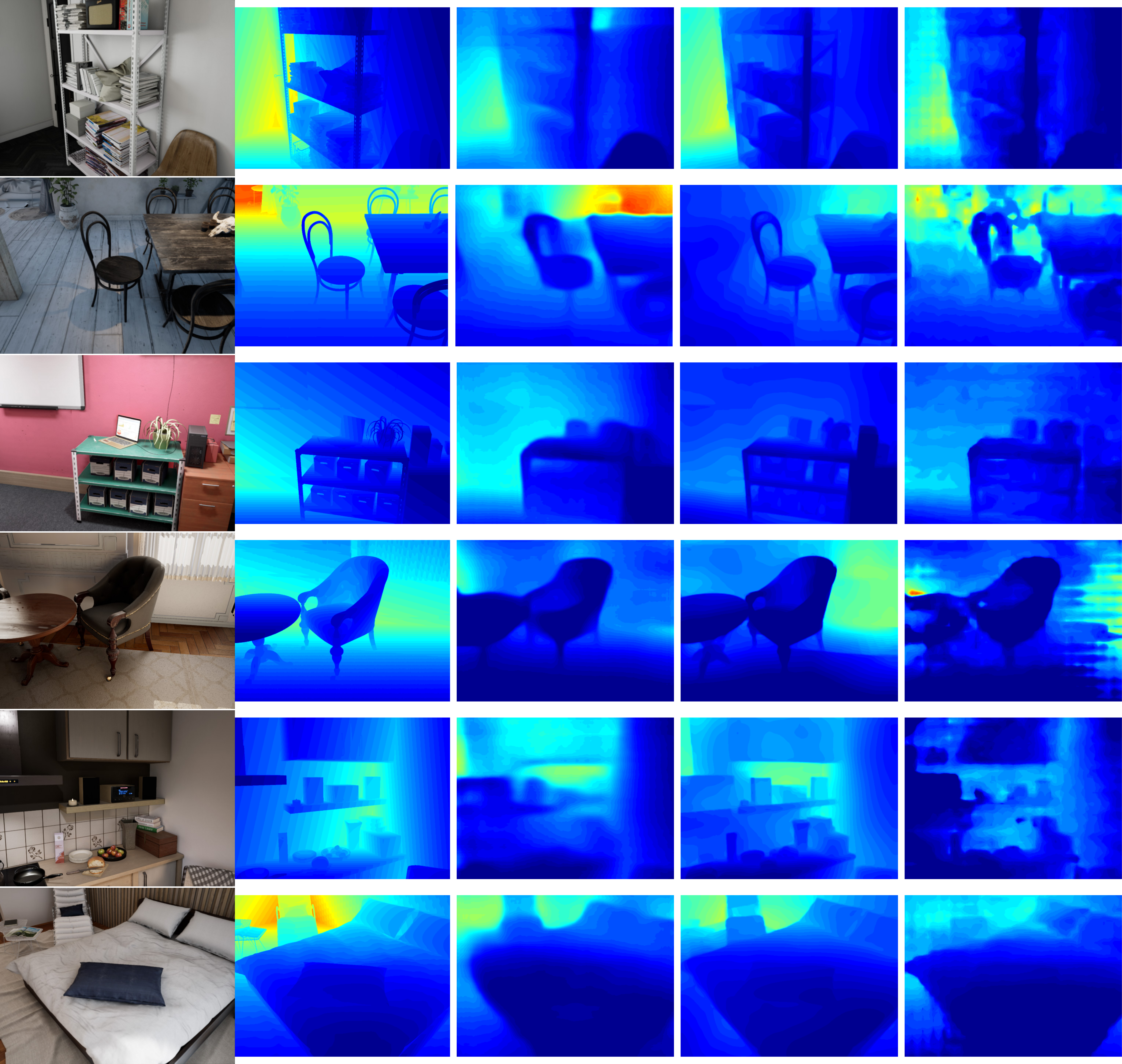}
\end{center}
   \caption{\textbf{Visual comparison of estimated depth maps on the Unreal-1k dataset:} input RGB images, ground truth depth maps, results using Laina et al. \cite{Laina2016}, our estimated depth maps, results of Fu et al. \cite{Fu2018DeepOR}.}
\label{fig:unreal}
\end{figure*}

Tab. \ref{tab:1} shows quantitative comparisons in terms of the average errors over the entire Unreal-1k dataset. As can be seen, our method outperforms the other two methods. We also compute the qualitative measure mSSIM described earlier. Fig. \ref{fig:unreal} presents a visual comparison of the different predicted depth maps against the ground truth.

\begin{table*}[t]
\centering
\begin{tabular}{l|lll|lll|l}
\toprule
Method         & $\delta_{1}\uparrow$ & $\delta_{2}\uparrow$ & $\delta_{3}\uparrow$ & rel$\downarrow$   & rms$\downarrow$   & $log_{10}\downarrow$ & mSSIM$\uparrow$  \\ 
\midrule
Laina et al. \cite{Laina2016}    & 0.526   &  0.786  & 0.896  & 0.311  & 1.049  &  0.130 & 0.903 \\
Fu et al. \cite{Fu2018DeepOR}    & \textbf{0.545} &  0.794  & 0.898  & 0.313  & 1.040  & 0.128 & 0.895 \\
\textbf{Ours}                    & 0.544  &  \textbf{0.803}  &  \textbf{0.904} & \textbf{0.301}  &  \textbf{1.030}  & \textbf{0.125} & \textbf{0.910}  \\
\bottomrule
\end{tabular}
\bigskip
\caption{\textbf{Comparisons of different methods on the Unreal-1k dataset.} Both the quantitative and qualitative metrics are presented. Note that even for the best performing methods the errors are still considerably large.}
\label{tab:3}
\end{table*}

\section{Conclusion}

In this work, we proposed a convolutional neural network for depth map estimation for single RGB images by leveraging recent advances in network architecture and the availability of high performance pre-trained models. We show that having a well constructed encoder, that is initialized with meaningful weights, can outperform state-of-the-art methods that rely on either expensive multistage depth estimation networks or require designing and combining multiple feature encoding layers. Our method achieves state-of-the-art performance on the NYU Depth v2 dataset and our proposed Unreal-1K dataset. Our aim in this work is to push towards generating higher quality depth maps that capture object boundaries more faithfully, and we have shown that this is indeed possible using an existing architectures. Following our simple architecture, one avenue for future work is to substitute the proposed encoder with a more compact one in order to enable quality depth map estimation on embedded devices. We believe their are still many possible cases of leveraging standard encoder-decoder models alongside transfer learning for high quality depth estimation. Many questions on the limits of our proposed network and identifying more clearly the effect on performance and contribution of different encoders, augmentations, and learning strategies are all interesting to purse for future work.

{ \small \bibliographystyle{zieee} \bibliography{zbib} }

\appendix

\section{Appendix}

\subsection{Network Architecture}

Tab. \ref{table:1} shows the structure of our encoder-decoder with skip connections network. Our encoder is based on the DenseNet-169 \cite{huang2017densely} network where we remove the top layers that are related to the original ImageNet classification task. For our decoder, we start with a $1 \times 1$ convolutional layer with the same number of output channels as the output of our truncated encoder. We then successively add upsampling blocks each composed of a $2\times$ bilinear upsampling followed by two $3 \times 3$ convolutional layers with output filters set to half the number of inputs filters, and were the first convolutional layer of the two is applied on the concatenation of the output of the previous layer and the pooling layer from the encoder having the same spatial dimension. Each upsampling block, except for the last one, is followed by a leaky ReLU activation function \cite{Maas13LeRELU} with parameter $\alpha=0.2$. The input images are represented by their original colors in the range $[0,1]$ without any input data normalization. Target depth maps are clipped to the range $[0.4, 10]$ in meters.

\begin{table}[t]
\centering
\resizebox{.45\textwidth}{!}{%
\begin{tabular}{ |l|l|l| } 
\hline
\textsc{Layer} & \textsc{Output} & \textsc{Function} \\
\hline
INPUT & $480 \times 640 \times 3$ &  \\ 
\hline
CONV1 & $240 \times 320 \times 64$ & DenseNet CONV1 \\ 
\hline
POOL1 & $120 \times 160 \times 64$ & DenseNet POOL1 \\ 
\hline
POOL2 & $60 \times 80 \times 128$ & DenseNet POOL2 \\ 
\hline
POOL3 & $30 \times 40 \times 256$ & DenseNet POOL3 \\ 
\hline
\ldots & \ldots & \ldots \\ 
\hline
CONV2 & $15 \times 20 \times 1664$ & \vtop{\hbox{\strut Convolution $1 \times 1$}\hbox{\strut of DenseNet BLOCK4}} \\ 
\hline
UP1 & $30 \times 40 \times 1664$ & Upsample $2 \times 2$ \\ 
\hline
CONCAT1 & $30 \times 40 \times 1920$ & Concatenate POOL3 \\ 
\hline
UP1-CONVA & $30 \times 40 \times 832$ & Convolution $3 \times 3$ \\ 
\hline
UP1-CONVB & $30 \times 40 \times 832$ & Convolution $3 \times 3$ \\ 
\hline
UP2 & $60 \times 80 \times 832$ & Upsample $2 \times 2$ \\ 
\hline
CONCAT2 & $60 \times 80 \times 960$ & Concatenate POOL2 \\ 
\hline
UP2-CONVA & $60 \times 80 \times 416$ & Convolution $3 \times 3$ \\ 
\hline
UP2-CONVB & $60 \times 80 \times 416$ & Convolution $3 \times 3$ \\ 
\hline
UP3 & $120 \times 160 \times 416$ & Upsample $2 \times 2$ \\ 
\hline
CONCAT3 & $120 \times 160 \times 480$ & Concatenate POOL1 \\ 
\hline
UP3-CONVA & $120 \times 160 \times 208$ & Convolution $3 \times 3$ \\ 
\hline
UP3-CONVB & $120 \times 160 \times 208$ & Convolution $3 \times 3$ \\ 
\hline
UP4 & $240 \times 320 \times 208$ & Upsample $2 \times 2$ \\ 
\hline
CONCAT3 & $240 \times 320 \times 272$ & Concatenate CONV1 \\ 
\hline
UP2-CONVA & $240 \times 320 \times 104$ & Convolution $3 \times 3$ \\ 
\hline
UP2-CONVB & $240 \times 320 \times 104$ & Convolution $3 \times 3$ \\ 
\hline
CONV3 & $240 \times 320 \times 1$ & Convolution $3 \times 3$ \\ 
\hline
\end{tabular}%
}
\bigskip
\caption{\textbf{Network architecture}. Layers up to CONV2 are exactly those of DenseNet-169 \cite{huang2017densely}. Upsampling is bilinear upsampling. We follow each CONVB convolutional layer by a leaky ReLU activation function \cite{Maas13LeRELU} with parameter $\alpha=0.2$. Note that in this table we use the output shapes corresponding to the spatial resolution of the dataset NYU Depth v2 ($height \times width \times channels$).}
\label{table:1}
\end{table}

\subsection{The Unreal-1K Dataset}

\begin{figure}[t]
\begin{center}
\includegraphics[width=\linewidth]{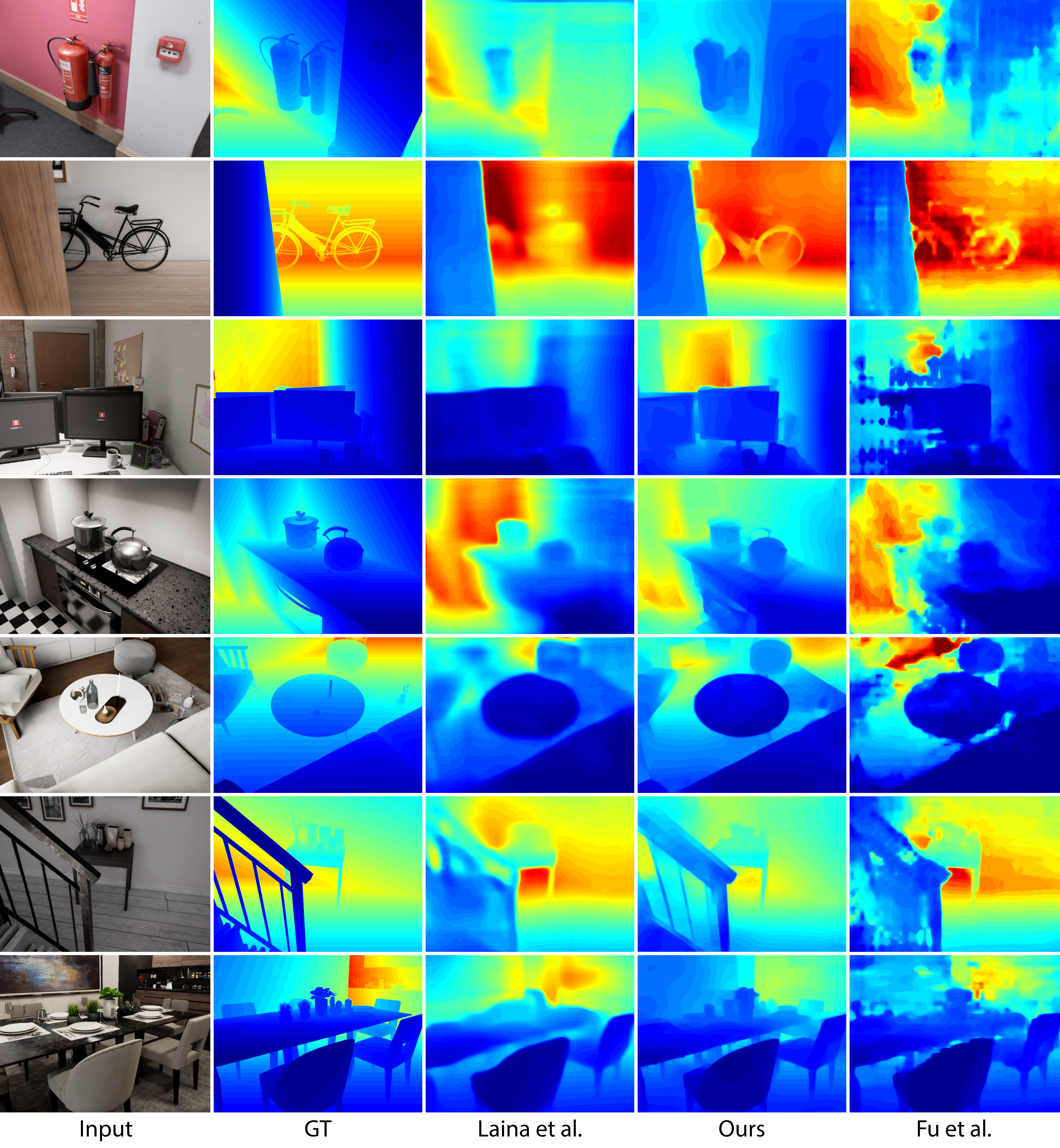}
\end{center}
   \caption{\textbf{Visual comparison of estimated depth maps on the Unreal-1K dataset:} input RGB images, ground truth depth maps, results using Laina et al. \cite{Laina2016}, our estimated depth maps, results of Fu et al. \cite{Fu2018DeepOR}. Note that, for better visualization, we normalize each depth map with respect to the range of its corresponding ground truth.}
\label{fig:append-fig1}
\end{figure}

We propose a new dataset of photo-realistic synthetic indoor scenes having near perfect ground truth depth maps. The scenes cover categories including living areas, kitchens, and offices all of which have realistic material and different lighting scenarios. These scenes, 32 scenes in total, are collected from the Unreal marketplace community \cite{UnrealMarket2018}. For each scene we select around 40 objects of interest and we fly a virtual camera around the object and capture images and their corresponding depth maps of resolution $640 \times 480$. In all, we collected more than 20K images from which we randomly choose 1K images as our testing dataset Unreal-1k. Fig. \ref{fig:append-fig1} shows example images from this dataset along with depth estimations using various methods.

\subsection{Additional Ablation Studies}
\label{sec:sup_ablation}

\begin{figure}[t!]
\begin{center}
\includegraphics[width=\linewidth]{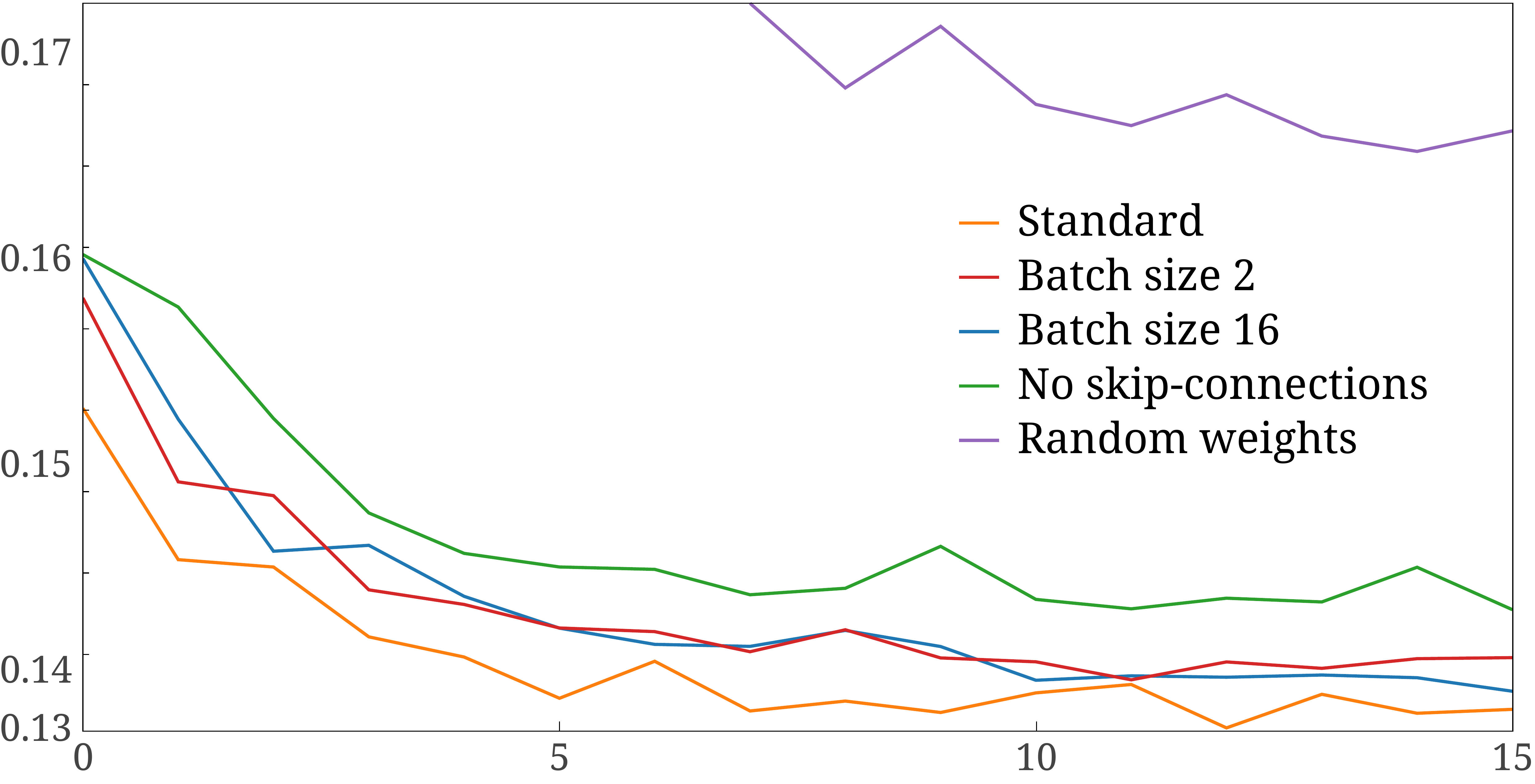}
\end{center}
   \caption{\textbf{Additional ablation studies.} Four variations on our standard model are considered. The horizontal axis represents the number of training iterations (in epochs). The vertical axis represents the average loss of the validation set at each epoch. Please see Sec. \ref{sec:sup_ablation} for more details. }
\label{fig:sup_ablation}
\end{figure}

We perform additional ablation studies to analyze more details of our proposed architecture. Fig. \ref{fig:sup_ablation} shows a representative look into the testing performance, in terms of validation loss, when changing some parts of our standard model. The training in these experiments is performed on the NYU Depth v2 dataset \cite{Silberman2012} for 750K iterations (15 epochs).

\paragraph{Pre-trained model.} In this experiment, we examine the effect of using an encoder that is initialized using random weights as opposed to being pre-trained on ImageNet which is what we use in our proposed standard model. In Fig. \ref{fig:sup_ablation} (purple), we can see the validation loss is greatly increased when training from scratch. This further validates that the performance of our depth estimation is positively impacted by transfer learning.

\paragraph{Skip connections.} In this experiment, we examine the effect of removing the skip connections between layers of the encoder and decoder. In Fig. \ref{fig:sup_ablation} (green), we can see the validation loss is decreased, compared to our proposed standard model, resulting in worse depth estimation performance.

\paragraph{Batch size.} In this experiment, we look at different values for the batch size and its effect on performance. In Fig. \ref{fig:sup_ablation} (red and blue), we can see the validation loss for batch sizes 2 and 16 compared to our standard model (orange) with batch size 8. Setting the batch size to 8 results in the best performance out of the three values while also training for a reasonable amount of time.

\end{document}